%% file: main.tex
\documentclass[10pt,twocolumn,letterpaper,pagenumbers]{article}

\usepackage{bm}
\usepackage[pagenumbers]{iccv} 
\usepackage{multirow}
\usepackage{makecell}
\usepackage{svg} 


\definecolor{iccvblue}{rgb}{0.21,0.49,0.74}
\usepackage[pagebackref,breaklinks,colorlinks,allcolors=iccvblue]{hyperref}

\def\paperID{1} 
\def\confName{ICCV}
\def\confYear{2025}

\title{LightGlueStick: a Fast and Robust Glue for Joint Point-Line Matching}

\author{
Aidyn Ubingazhibov$^{1}$ \quad
Rémi Pautrat$^{2}$ \quad
Iago Suárez$^{3}$ \quad
Shaohui Liu$^{1}$ \\
Marc Pollefeys$^{1,2}$ \quad
Viktor Larsson$^{4}$ \\
$^{1}$ETH Zurich \quad
$^{2}$Microsoft Spatial AI Lab \quad
$^{3}$Qualcomm XR Labs Europe \quad
$^{4}$Lund University \\
}

\begin{document}
\maketitle
\input{sec/0_abstract}
\input{sec/1_intro.tex}

\input{sec/2_related_work.tex}
\input{sec/3_Method}
\input{sec/4_Experiments}
\input{sec/5_Conclusion}

{
    \small
    \bibliographystyle{ieeenat_fullname}
    \bibliography{main}
}


\end{document}

%% file: sec/0_abstract.tex
\begin{abstract}
Lines and points are complementary local features, whose combination has proven effective for applications such as SLAM and Structure-from-Motion. The backbone of these pipelines are the local feature matchers, establishing correspondences across images. Traditionally, point and line matching have been treated as independent tasks. Recently, GlueStick proposed a GNN-based network that simultaneously operates on points and lines to establish matches. While running a single joint matching reduced the overall computational complexity, the heavy architecture prevented real-time applications or deployment to edge devices.

Inspired by recent progress in point matching, we propose LightGlueStick, a lightweight matcher for points and line segments. The key novel component in our architecture is the  Attentional Line Message Passing (ALMP), which explicitly exposes the connectivity of the lines to the network, allowing for efficient communication between nodes. In thorough experiments we show that LightGlueStick establishes a new state-of-the-art across different benchmarks. The code is available at \href{https://github.com/aubingazhib/LightGlueStick}{\url{https://github.com/aubingazhib/LightGlueStick}}.

\end{abstract}

%% file: sec/1_intro.tex
\section{Introduction}
Visual feature matching is a fundamental component in numerous computer vision tasks, including visual localization, pattern detection, Structure-from-Motion (SfM), and Simultaneous Localization and Mapping (SLAM). These tasks are essential for navigation and positioning in fields such as robotics, augmented reality, architecture, and manufacturing. Classical handcrafted keypoint detectors and descriptors powered some of the first advances in these areas. However, these methods typically require well-textured regions and struggle with viewpoint and illumination changes. Conversely, line segments are abundant in texture-less areas, provide robust structural cues, and are resilient against occlusion and illumination changes. Their complementary nature makes the joint matching of points and lines an interesting paradigm for robust sparse image matching.

\begin{figure}[t]
    \centering
    \includesvg[width=\linewidth]{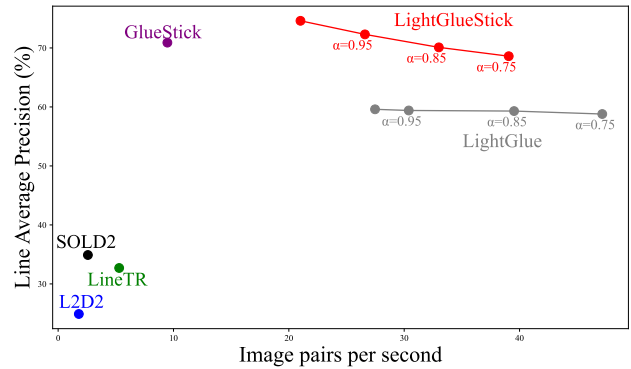}
    \caption{{\bf LightGlueStick outperforms other state-of-the-art line matchers with a speed comparable to LightGlue~\cite{lindenberger2023lightglue}}. The depth adaptivity mechanism allows to predict line matches earlier. The parameter $\alpha$ controls the tradeoff between speed and accuracy. Here, LightGlue~\cite{lindenberger2023lightglue} matches lines using endpoints.}
    \label{fig:example}
\end{figure}

However, line matching presents unique challenges. Line segments are frequently partially occluded or fragmented, and in general less repeatable compared to points, at least in terms of endpoint consistency.
As each line can span a large region of the image, their local descriptors are also more affected by foreshortening and other perspective effects.
This yields a higher sensitivity to viewpoint changes, making them harder to both detect and describe.
This is further complicated by lines frequently appearing in texture-poor areas, e.g.~delineating walls or other planar surfaces.

Moreover, many image-matching applications require on-device execution with strict latency and power constraints. These conditions limit the practicality of recent dense~\cite{sun_2021_loftr,edstedt2023dkm,edstedt2024roma} and semi-dense~\cite{jiang_2021_cotr} matching methods, which typically demand significant computational resources. Consequently, sparse matching techniques become preferable due to their fast execution with minimal hardware requirements. A lightweight scene representation is also a requirement for real-time mapping purposes when the map needs to be compact or efficiently transmitted over limited-bandwidth networks.

Recently, deep learning methods have significantly outperformed classical matching and filtering techniques, such as mutual nearest neighbor and ratio test~\cite{lowe_2004}. Their key to success is to relate not only to the visual descriptors representing the image appearance but also to consider the local feature geometry, i.e.~how the points and lines are spatially distributed in the images.
These learned matchers have since replaced classical hand-crafted heuristics based on local planarity, coherence, or filtering~\cite{wang2009wide,fan_2012_robust,ramalingam_2015_line,ma2019locality}. 
HDPL~\cite{guo_2021_hdpl} and GlueStick~\cite{pautrat_suarez_2023_gluestick} introduced the first learning approaches for joint matching of points and lines, yet the computational cost limits their applicability in real-time, on-device scenarios.
Recently, LightGlue~\cite{lindenberger2023lightglue} has gained attention for significantly accelerating point-only matching through efficient attention mechanisms and reduced redundant computations.

In this paper, we introduce the next generation of point-line matching with LightGlueStick, a novel matcher that marries the matching capabilities of GlueStick with the computational efficiency of LightGlue.
This facilitates effective joint matching of points and lines, making it suitable for real-time and on-device applications.
Considering the importance of speed for downstream tasks, our approach ensures that joint matching remains practical by significantly reducing inference latency. 
As we show in our experiments, LightGlueStick surpasses current state-of-the-art sparse methods in both accuracy and runtime performance.

We re-design the joint point-line matching strategy with optimized attention mechanisms~\cite{rotary_enc} into a unified architecture.
Our method represents keypoints and line endpoints as nodes within a graph-based structure, where edges dynamically adapt across layers.
We leverage self-attention and cross-attention techniques from transformers~\cite{vaswani_2017_transformers,sarlin_2020_superglue}, complemented by our novel Attentional Line Message Passing (ALMP) layer. 
This layer explicitly encodes line connectivity, enhancing communication between endpoints while allowing the network to ignore non-repeatable line endpoints. 
Extensive evaluation across multiple benchmarks demonstrates that LightGlueStick achieves state-of-the-art performance in multiple tasks. The code will be released upon publication. Our key contributions are:
\begin{itemize}
\item We achieve significant speedups while maintaining superior performance compared to state-of-the-art point and line matchers.
\item We introduce a new Attentional Line Message Passing layer to improve the communication between line endpoints.
\item Our approach is flexible, with a tunable threshold to trade off matching accuracy and efficiency.
\item Extensive experiments demonstrate that our approach sets new state-of-the-art results across multiple benchmarks.
\end{itemize}


%% file: sec/2_related_work.tex
\section{Related Work}

{\bf Point matching.} 
Feature point matching has a long history in computer vision, and was originally performed by extracting handcrafted descriptors from image patches~\cite{lowe_2004,bay2008,orb}. Point matches were then obtained with either mutual nearest neighbor or ratio test~\cite{lowe_2004}. The resurgence of deep learning introduced a series of more robust descriptors by extracting learned embeddings of image patches~\cite{yi2016,yurun2017,mishchuk2017,ono2018lfnet,sosnet2019cvpr,Luo2019}. Later, feature descriptors started to be extracted densely~\cite{detone2018,revaud2019,Dusmanu2019CVPR,tyszkiewicz2020disk,Zhao2022ALIKE,Zhao2023ALIKED,silk,edstedt2024dedode,suarez2021revisiting,potje2024cvpr}, so that point features can be retrieved with bi-linear interpolation from the dense map. Learned matchers were originally introduced with the pioneering work SuperGlue~\cite{sarlin_2020_superglue}, which proposed to combine Graph Neural Networks (GNNs) and transformers to enrich the point descriptors with the context of the other points. Subsequent works improved the original architecture to make it less costly to use~\cite{chen2021sgmnet,Shi2022ClusterGNNCC,lindenberger2023lightglue}. LightGlue~\cite{lindenberger2023lightglue} proposed in particular an adaptivity mechanism allowing to finish the inference earlier for easy image pairs, thus significantly enhancing efficiency. OmniGlue~\cite{jiang2024Omniglue} improved the robustness of the matching even further, by incorporating features from a vision foundation model. Instead of matching sparse keypoints, another line of research aims at matching image pairs densely. Initiated by LoFTR~\cite{sun_2021_loftr}, this field has been very active in recent years~\cite{sun_2021_loftr, truong_2021_learning,jiang_2021_cotr,wang2022matchformer,Chen2022ASpanFormerDI,edstedt2023dkm,edstedt2024roma,mast3r_arxiv24,he2025matchanything}. These matches obtain good results at a higher computational cost.
In this work, we take inspiration from the breakthroughs in efficiency brought by LightGlue~\cite{lindenberger2023lightglue}, and incorporate them into a sparse joint point-line matcher.

\begin{figure*}[t]
    \centering
    \includesvg[width=\textwidth]{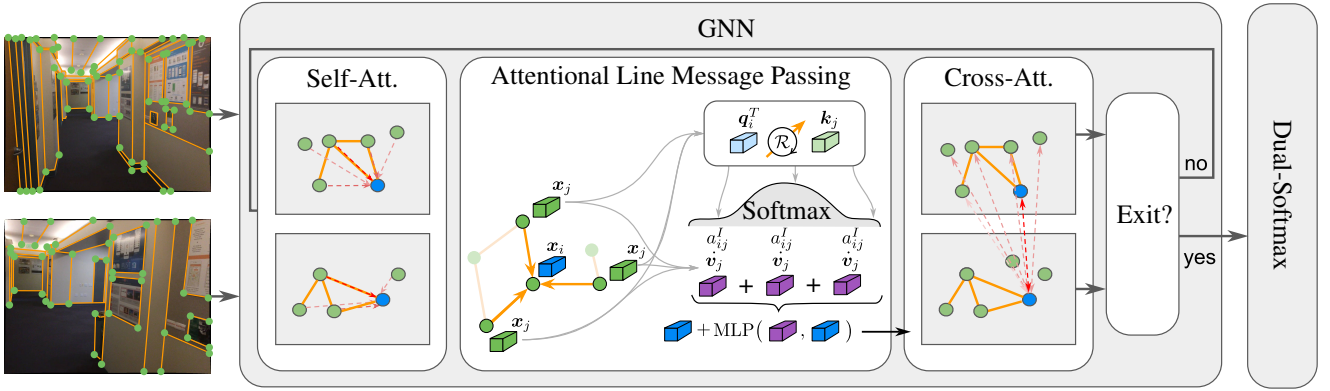}
    \caption{\textbf{Overview of the proposed pipeline}: LightGlueStick takes as input two images where points and lines have been detected and described. Our Graph Neural Network (GNN) is composed of three stages: 1) Self-Attention, 2) our novel Attentional Line Message Passing, and 3) Cross-Attention. Depending on the achieved matching result at each layer, we can stop the execution to save computation. The last step of Dual-Softmax generates two assignation matrices for points and lines, respectively.}
    \label{fig:pipeline}
\end{figure*}

{\bf Line matching.} 
Similarly as for points, line matching has originally been performed by extracting image patches around lines and by computing handcrafted heuristics to describe the image gradient~\cite{wang2009msld,wang2009wide,zhang_2013_lbd,verhagen2014scale,li_2016_ljl}. In the same spirit, early line matchers followed the same trend by running neural networks on the image patches around lines~\cite{lange_2019_dld,Kruger_2020_wld,l2d2}, bringing additional robustness to the line descriptors. Further improvements of the learned line descriptors involved extracting dense line descriptors~\cite{vakhitov_2019_lld,pautrat_2021_sold2}, sampling and describing points along the lines to increase the robustness to occlusions~\cite{l2d2,pautrat_2021_sold2}, and jointly extracting lines and their descriptors with a single network~\cite{l2d2,pautrat_2021_sold2}. With the advent of transformers and GNNs, recent methods used attention within a line to enrich its line descriptor~\cite{syoon_2021_linetr}, while others connected the line segments into a graph and added self- and cross-attention between line nodes~\cite{ma_2021_wglsm,guo_2021_hdpl,pautrat_suarez_2023_gluestick}. While WGLSM~\cite{ma_2021_wglsm} only used line features, HDPL~\cite{guo_2021_hdpl} and GlueStick~\cite{pautrat_suarez_2023_gluestick} combined points and lines within the same graph. The former represents lines as a single node in the graph, while the latter leverages the line endpoint descriptors to ease communication between point and line features. Although GlueStick~\cite{pautrat_suarez_2023_gluestick} is the current state-of-the-art point and line matcher, its execution time remains slow, making it unpracticable for real-time applications. Our work improves the architecture of GlueStick, by making it more efficient, while maintaining or improving its performance.

{\bf Joint use of points and lines.}
Jointly matching point and line features unleashes multiple applications fusing the two kinds of features. SLAM has a long history of leveraging both points and line segments~\cite{zuo2017robust,pumarola2017pl,gomez2019pl,fu2020plvins,lim2022uv,pljslam,eplf-vins,Li2023IDLSID}. The long range of lines is indeed a strong asset to reliably track features in a sequence and to reduce the drift over time. Line by-products can also be used in SLAM to add constraints, such as their junctions~\cite{pljslam} and the vanishing points~\cite{suarez2018fsg, lim2022uv}. \citet{Liu_2023_LIMAP} proposed to reconstruct lines in 3D, based on an existing point reconstruction, and further extended it into a full SfM pipeline integrating points, lines, and vanishing points~\cite{Liu2024RobustIS}. Jointly using point and line features can also benefit visual localization and the related solvers by providing additional constraints, as demonstrated in~\cite{ramalingam2011pose,vakhitov2016accurate,zhou2018stable,agostinho2019cvxpnpl,zhou2020fast,Liu_2023_LIMAP}. Finally, while lines do not bring constraints for relative pose estimation, \cite{Hrub2023HandbookOL} showed that one could combine points, line endpoints, and vanishing points to better constrain the relative pose between two images.

%% file: sec/3_Method.tex
\section{LightGlueStick}


Our goal in this work is to match a given set of point and line features between two images $A$ and $B$. We use $I$ to refer generically to either image. We adopt the same wireframe formulation as in \cite{pautrat_suarez_2023_gluestick}, where we represent lines with their two endpoints and initialize a graph where nodes consist of keypoints and line endpoints, and edges connect endpoints that are part of the same line. We first extract a dense descriptor map of dimension D for the entire image using an off-the-shelf network, from which we can interpolate the initial embedding $\bm{x}_i^I \in \mathbb{R}^{D}$ of each node $i$ in image $I$. We describe in \cref{sec:gnn_backbone} how we enrich these initial node embeddings with the visual and spatial context of the other nodes, through a Graph Neural Network (GNN)~\cite{Bronstein2016GeometricDL} and successive layers of transformer operations~\cite{vaswani_2017_transformers}. After the GNN, we leverage the enriched node embeddings to associate the point and line features of both images (\cref{sec:assignment}). The overview of our method is available in \cref{fig:pipeline}.

\subsection{Transformer Backbone}
\label{sec:gnn_backbone}

We describe in the following how node features are enriched in our proposed GNN, which consists of a series of 3 layers: self-attention, cross-attention, and a novel Attention Line Message Passing (ALMP). Each of these three layers is repeated up to $L$ times to successively enrich the representation of each node before establishing matches.

{\bf Residual feature update}. In each layer, the feature vector $\bm{x}_i^{I}$ of a node $i$ is updated by applying a Multi-Layer Perceptron (MLP) to the concatenation of $\bm{x}_i^{I}$ and $\bm{m}_i^{I \leftarrow S}$, where $\bm{m}_i^{I \leftarrow S}$ represents a message received from a source image $S \in \{A, B\}$: 
\begin{equation}
\bm{x}_i^{I} \gets \bm{x}_i^{I} + \text{MLP}\big(\big[\bm{x}_i^{I} | \bm{m}_i^{I \leftarrow S}\big]\big),
\label{eq:message_passing}
\end{equation}
where $[\cdot | \cdot]$ represents concatenation. In self-attention layers, messages are aggregated only from nodes within the same image, implying \(S = I\). In cross-attention layers, node states are updated using nodes from the other image, meaning \(S = \{A, B\} \setminus I\).

{\bf Self- and cross-attention layers.}
We use a lightweight attention mechanism~\cite{rotary_enc, lindenberger2023lightglue} for the implementation of the self- and cross-attention.
The message is computed by an attention mechanism:
\begin{equation}
\bm{m}_i^{I \leftarrow S} = \sum_{j \in S} \text{Softmax}_{k \in S}(a_{ik}^{IS})_{j} \bm{v}_j^{S},
\end{equation}
where $a_{ij}^{IS}$ is an attention weight between point $i \in I$ and point $j \in S$, while $\bm{v}_{j}^{S} \in \mathbb{R}^{D}$ is the value vector obtained by applying a linear transformation to $\bm{x}_{j}^{S}$. We define next how this attention is computed.

Consider nodes \( i \) and \( j \) at positions \(\bm{p}_i, \bm{p}_j \in \mathbb{R}^2\). For self layers, we define the attention score as
\begin{equation}
a_{ij}^{I} = \bm{q}_i^T\, \mathcal{R}(\bm{p}_j - \bm{p}_i)\, \bm{k}_j,   
\label{eq:attention_formula}
\end{equation}
where \(\bm{q}_i, \bm{k}_j \in \mathbb{R}^K\) are obtained from \(\bm{x}_i^I\) and \(\bm{x}_j^I\) via separated linear transforms. Here, \(\mathcal{R}(\bm{d})\) is a rotary positional encoding~\cite{rotary_enc} that increases attention between nearby elements in a learned distance space. To ensure translation equivariance, we use the relative position \(\bm{d} = \bm{p}_j - \bm{p}_i\) and project it onto learned 2D vectors \(\bm{b}_k\) to obtain \(K/2\) angles \(\theta_k\). This defines
\begin{equation}
    \*\mathcal{R}\!\left(\*\bm{d}\right) = 
    \left(
    \begin{smallmatrix}
    \hat{\*\mathcal{R}}(\*\bm{b}_1^\top\*\bm{d}) && \*0\\
    &\ddots\\
    \*0 && \hat{\*\mathcal{R}}(\*\bm{b}_{K\!/\!2}^\top\*\bm{d})
    \end{smallmatrix}
    \right)
    \!, 
    \hat{\*\mathcal{R}}(\theta_k) = \left(
    \begin{smallmatrix}
    \cos\theta_k&-\sin\theta_k\\
    \sin\theta_k&\cos\theta_k
    \end{smallmatrix}
    \right).
\end{equation}
Thus, \(\mathcal{R}(\bm{d})\) splits the \(K\)-dimensional space into \(K/2\) 2D subspaces and rotates each by \(\theta_k\) (akin to Fourier Features~\cite{rahimi2007random, fourierfeatures}), combining spatial relative location with the query-key alignment in \(a_{ij}^{I}\). In our implementation, we use $K=64$ with four parallel attention heads.



For cross layers, the attention score $a_{ij}^{IS}$ is computed as
\begin{equation}
a_{ij}^{IS} = (\bm{k}_i^{I})^\top \, \bm{k}_j^{S},
\end{equation}
where the cross layers are bidirectional to save computation.

{\bf Attentional Line Message Passing}. GlueStick introduced the Line Message Passing (LMP) layer, which averages messages from neighboring nodes based on line connectivity. However, line detectors can produce non-repeatable lines across views, causing inconsistencies in line connectivity for some endpoints that are otherwise matchable. To account for this, we propose Attentional Line Message Passing (ALMP), which aggregates messages with an attention mechanism.

Line endpoint feature vectors are updated as in \cref{eq:message_passing}, where the messages are passed between nodes of the same image. The message is computed as:
\begin{equation}
\bm{m}_i^{I} = \sum_{j \in \mathcal{N}(i) \cup \{i\}} \text{Softmax}_{k \in \mathcal{N}(i) \cup \{i\}}(a_{ik})_{j} \,\bm{v}_j^I,
\end{equation}
where $\mathcal{N}(i)$ denotes the set of neighboring endpoint nodes of $i$ based on its line connectivity, that is, endpoints directly connected to $i$ by wireframe construction. The term $a_{ik}$ represents the attention score between endpoints $i$ and $k$, calculated as in \cref{eq:attention_formula}, and $\bm{v}_{j}^{I} \in \mathbb{R}^{D}$ is a value vector obtained by applying a linear transformation to the endpoint feature vector $x_{j}^{I}$.\\
By rotating feature vectors based on the relative positions of line endpoints, the network effectively captures the angular relationships between lines meeting at a junction, thereby facilitating the matching of junctions with similar angular connectivity across views. The learned attention further enables the network, after the cross-layer, to identify endpoints that are non-repeatable but connected to repeatable ones, and to ignore them for improved matching.

\subsection{Correspondence Prediction}
\label{sec:assignment}
{\bf Points.} To predict point correspondences, we compute an assignment matrix based on point-wise similarity scores. The similarity score $s_{ij}^{p}$ for points $i$ in $A$ and $j$ in $B$ is:
\begin{equation}
s_{ij}^{p} = \text{Linear}(\bm{x}_i^{A})^T \text{Linear}(\bm{x}_j^{B}),
\end{equation}
where $\text{Linear}(\cdot)$ is a learned linear transformation.
We compute a matchability score for each point $i \in I$ as
\begin{equation}
\sigma_{i}^{p,I} = \text{Sigmoid}\big(\text{Linear}(\bm{x}_i^I)\big).
\label{eq:matchability}
\end{equation}

Finally, the assignment matrix $S_{ij}^{p}$ combines the point similarity and matchability:
\begin{equation}
\bm{S}_{ij}^{p} = \sigma_i^{p, A} \sigma_j^{p, B} \text{Softmax}_{j \in B}(s_{ij}^{p}) \text{Softmax}_{i \in A}(s_{ij}^{p}).
\end{equation}

{\bf Lines.} Lines are matched similarly, but by matching endpoints in an order-agnostic manner. Suppose $x_{s}^{I}$, $x_{e}^{I}$ are feature vectors of two endpoints of the same line in image $I$. We first project the features with a linear transform:
\begin{equation}
    \bm{y}_s^I = \text{Linear}(\bm{x}_s^I),\quad \bm{y}_e^I = \text{Linear}(\bm{x}_e^I)
\end{equation}
The similarity score $s_{ij}^{l}$ for lines $i$ in $A$ and $j$ in $B$ is then:
\begin{equation}
\begin{aligned}
s_{ij}^{l} = \max\Big( & (\bm{y}_{s}^{A})^{T}\bm{y}_{s}^{B} + (\bm{y}_{e}^{A})^{T}\bm{y}_{e}^{B}, \\
                        & (\bm{y}_{s}^{A})^{T}\bm{y}_{e}^{B} + (\bm{y}_{e}^{A})^{T}\bm{y}_{s}^{B} \Big).
\end{aligned}
\end{equation}
We define the matchability score of an endpoint as in Equation~\eqref{eq:matchability}, but with a distinct linear layer. This ensures that the noise from matching less repeatable endpoints does not affect the point matching. The matchability score \(\sigma_i^{l}\) for line \( i \) is then computed as the average of its endpoint matchabilities:
\begin{equation}
\sigma_i^{l} = \frac{\text{Sigmoid}\big(\text{Linear}(\bm{x}_s)\big) + \text{Sigmoid}\big(\text{Linear}(\bm{x}_e)\big)}{2},
\end{equation}
where $x_{s}$ and $x_{e}$ are the endpoint descriptors.
Similar to keypoints, the assignment matrix for lines $S_{ij}^{l}$ combines line similarity and matchability:
\begin{equation}
\bm{S}_{ij}^{l} = \sigma_{i}^{l,A} \sigma_{j}^{l,B} \text{Softmax}_{j \in B}(s_{ij}^{l}) \text{Softmax}_{i \in A}(s_{ij}^{l})
\end{equation}

{\bf Match filtering}. A pair of points or lines (\(i\), \(j\)) is considered a valid correspondence if both points or lines are predicted to be matchable and if their similarity is higher than that of any other point or line in either image. We select pairs where $S_{ij}$ exceeds a threshold $\tau$ and is greater than all other elements in its respective row and column.

\subsection{Supervision}

We supervise our model using ground truth point and line matches derived from homographies or computed based on pixel-wise depth and relative pose, following the ground truth generation method of GlueStick~\cite{pautrat_suarez_2023_gluestick}. This results in the following ground truth matches: a set of positive matches, represented by index pairs $\mathcal{M}^p$ for points and $\mathcal{M}^{l}$ for lines, as well as sets of unmatchable point and line indices for both views, denoted as $\Tilde{A}^{p}$, $\Tilde{B}^{p}$ for points and $\Tilde{A}^{l}$, $\Tilde{B}^{l}$ for lines.

 To encourage early prediction of correspondences, the model is supervised after each layer. Dropping the index for points or lines, we define the negative log-likelihood of the predicted assignments for a given feature (points or lines) at layer $\ell$:
 \begin{equation}
\begin{aligned}
\mathcal{L}^{\ell}(\bm{S}_{ij}, \mathcal{M}, \Tilde{A}, \Tilde{B}) =
    & -\frac{1}{|\mathcal{M}|} \sum_{(i, j) \in \mathcal{M}} \log ^{\ell}\bm{S}_{ij} \\
    & - \frac{1}{2|\Tilde{A}|} \sum_{i \in \Tilde{A}} \log \left( 1 - ^{\ell}\sigma_i^{A} \right) \\
    & - \frac{1}{2|\Tilde{B}|} \sum_{j \in \Tilde{B}} \log \left( 1 - ^{\ell}\sigma_j^{B} \right).
\end{aligned}
\end{equation}
The final loss is an average of the point and line losses at all layers, where $L$ is the total number of layers:
\begin{equation}
\begin{aligned}
\mathcal{L} = \frac{\sum_{\ell} \mathcal{L}^{\ell}(\bm{S}_{ij}^{p}, \mathcal{M}{\strut^p}, \Tilde{A}{\strut^p}, \Tilde{B}{\strut^p}) + \mathcal{L}^{\ell}(\bm{S}_{ij}^{l}, \mathcal{M}{\strut^l}, \Tilde{A}{\strut^l}, \Tilde{B}{\strut^l})}{2L}.
\end{aligned}
\end{equation}

\subsection{Adaptive Depth and Width}
We adopt adaptivity mechanisms from LightGlue, specifically reducing the number of layers based on the difficulty of image pairs and pruning keypoints that are confidently predicted as unmatchable.\\
{\bf Confidence Classifier:} If an image pair is easy (i.e., has large visual overlap and minimal appearance shift), early-layer predictions will closely match those of later layers, so that we can stop the prediction of the network early. After each block $\ell$, we estimate the confidence that each node's assignment remains unchanged in the final output:
\begin{equation}
c_{i}^\ell = \text{Sigmoid}(\text{MLP}(\bm{x}_{i})).
\end{equation}
{\bf Exit criterion:} Following LightGlue, we terminate inference early when a sufficient fraction of nodes are confident. A node is considered confident if its predicted confidence exceeds a threshold $\lambda_{\ell}$ at layer $\ell$. Assuming $N$ nodes in image $A$ and $M$ nodes in image $B$, inference is halted if the fraction of confident nodes surpasses a predefined threshold $\alpha$:
\begin{equation}
    \text{exit} = \left( \frac{1}{N+M} \sum_{I \in \{A, B\}} \sum_{i \in I} \mathbb{I}[c_i^\ell > \lambda_\ell] \right) > \alpha.
\end{equation}
Note that we do not compute confidence scores for lines directly; instead, we rely on keypoint and endpoint confidences to obtain reliable line matches from earlier layers.\\
{\bf Pruning}.
When testing the point and line pruning proposed in LightGlue~\cite{lindenberger2023lightglue}, we did not observe a significant speedup brought by pruning strategies. In particular for lines, given that images often contain much fewer lines than points, the overhead brought by the line pruning is often surpassing the gains in speedup. Therefore, we did not further explore the pruning of nodes during inference.

%% file: sec/4_Experiments.tex
\section{Experiments}
We pre-train LightGlueStick on image pairs obtained by synthetic homography warps, sampled from the 1M distractor images of \cite{radenovic2018}. Following this, we fine-tune the model on the Megadepth dataset~\cite{Li_2018_MegaDepth}, which contains 196 scenes of tourist landmarks. We use gradient checkpointing \cite{Chen2016TrainingDN} to fit a batch of 32 image pairs on a single 40Gb Nvidia A100 GPU. The images are resized to 1024x1024 px. We used Adam optimizer with a learning rate of $10^{-4}$. Training is conducted on homographies for 40 epochs, followed by fine-tuning on MegaDepth for an additional 50 epochs. The learning rate is exponentially decayed after the 20th and 30th epochs, respectively. The entire training process took approximately 9 days.

{\bf Implementation details.} LightGlueStick has $L = 9$ blocks of self-attention, Attentional Line Message Passing (ALMP), and cross-attention layers. Each attention unit has 4 heads. The network is trained with 1500 keypoints and 250 lines, with nodes represented by $D=256$ dimensional features. We train LightGlueStick with Superpoint~\cite{detone2018} local features and use LSD~\cite{gioi_2010_lsd} to detect lines.

{\bf Baselines.} 
We compare LightGlueStick with the SOTA sparse point matcher LightGlue~\cite{lindenberger2023lightglue}; the dense matchers LoFTR~\cite{sun_2021_loftr} and RoMa~\cite{edstedt2024roma}; the line matchers: SOLD$^{2}$~\cite{pautrat_2021_sold2}, LineTR~\cite{syoon_2021_linetr}, and L2D2~\cite{l2d2}; and two methods combining points and lines: PL-Loc~\cite{syoon_2021_linetr} and GlueStick~\cite{pautrat_suarez_2023_gluestick}. SOLD$^{2}$~\cite{pautrat_2021_sold2} employs its own line detector, while all other methods including LightGlueStick use LSD~\cite{gioi_2010_lsd} for line detection, if not otherwise specified. ``LG+Endpoints'' refers to matching lines by first matching their endpoints with LightGlue~\cite{lindenberger2023lightglue}, then finding the best association of pairs of endpoints as in \cite{pautrat_suarez_2023_gluestick}.

\begin{table}[h]
    \centering
    \small
    \begin{tabular}{lcc}
        \toprule
        \textbf{Architecture} & \textbf{Line AP} & \textbf{Time (ms)} \\
        \midrule
        GlueStick & 70.9 & 105.6 \\
        \midrule
        {\bf LightGlueStick (ALMP)} & {\bf 74.6} & 47\\
        \quad $\hookrightarrow$ a) no LMP & 68.4 & {\bf 39} \\
        \quad $\hookrightarrow$ b) mean LMP & 73.3 & 54 \\
        \bottomrule
    \end{tabular}
    \caption{\textbf{Ablation study on the ETH3D dataset~\cite{Schops_2017_eth3d}.} We compare our ALMP against no LMP and GlueStick's mean LMP.}
    \label{tab:ablation}
\end{table}

\subsection{Ablation Study}
We evaluate the improvements brought by our contributions on the ETH3D dataset~\cite{Schops_2017_eth3d}. Since our improvements are not touching point matching, we show only the performance of line matching here. We use the images of the 13 training scenes of the high-resolution multi-view dataset, downsample them by a factor of 8, and sample pairs of images with at least 500 keypoints in common in the official reconstruction. The ground truth line matches are obtained by leveraging the dataset LiDAR depth and poses, using the same methodology as was used to supervise our network. Given predicted line matches, we rank them by matching score, and compute a precision-recall curve. We report the Average Precision (AP) in \cref{tab:ablation}, computed as the Area Under the Curve (AUC) of the precision-recall curve.

We compare the original GlueStick architecture~\cite{pautrat_suarez_2023_gluestick} with our final model (LightGlueStick (ALMP)), and also ablate the latter with either no LMP, or the same LMP as was used in GlueStick (referred to ``mean LMP'').
Starting from the original GlueStick, we can observe that incorporating the learnings from LightGlue~\cite{lindenberger2023lightglue} for line matching (rotary encoding, bi-directional attention, flash attention, matchability prediction, shown in line ``b) mean LMP'') already brings a significant boost of performance and efficiency. Using no LMP incurs however a severe drop of performance. On the contrary, our final model equipped with our proposed ALMP gives the best balance between performance and execution time.

\subsection{Line Matching Evaluation on ETH3D}

\begin{figure}[t]
    \centering
    \small
    \begin{minipage}{0.43\textwidth}
        \rotatebox{90}{(a) GlueStick~\cite{pautrat_suarez_2023_gluestick}}
        \includegraphics[width=\textwidth]{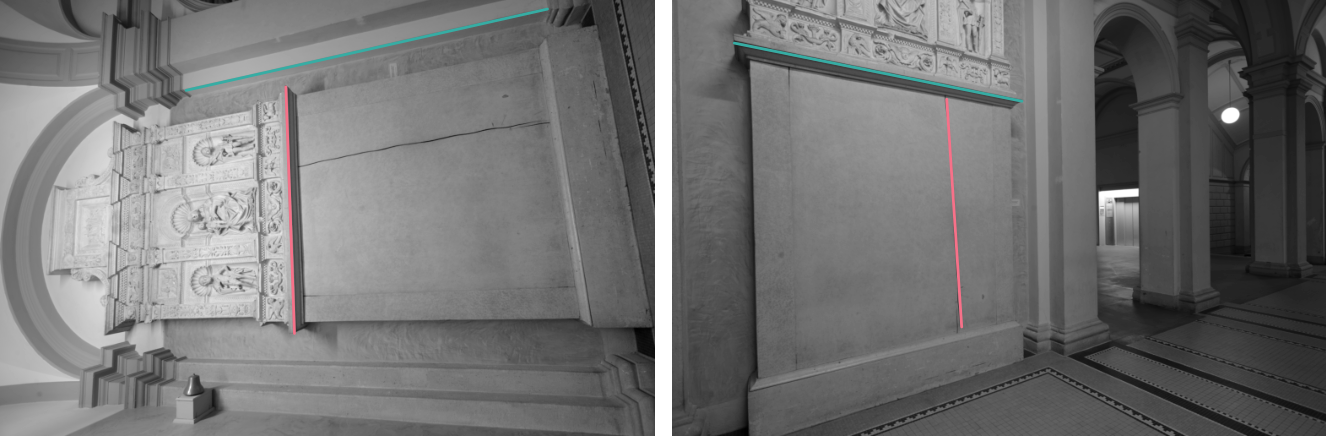}
    \end{minipage}\hspace{2em}
    \begin{minipage}{0.43\textwidth}
        \rotatebox{90}{\phantom{xxxx}(b) Ours}
        \includegraphics[width=\textwidth]{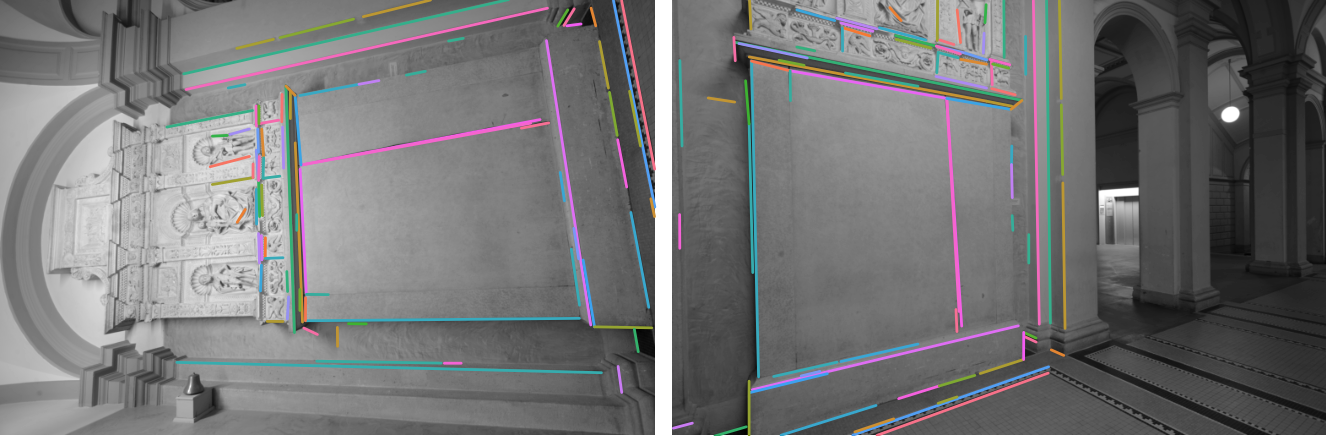}
    \end{minipage}
    \caption{LightGlueStick successfully predicts matches on rotated images, despite not being trained on them, whereas GlueStick~\cite{pautrat_suarez_2023_gluestick} fails to predict correct line matches.}
    \label{fig:rotated_example}
\end{figure}

We evaluate the quality of our point and line matches on the ETH3D dataset~\cite{Schops_2017_eth3d} by computing the Average Precision (AP) of either point or line matches. We also report the average running time per image pair for each method.
The results can be found in \cref{tab:eth3d_evaluation}. LightGlueStick obtains the best performance across the board. One can note in particular that it surpasses the previous best point-line matcher GlueStick~\cite{pautrat_suarez_2023_gluestick} on both points and lines, while being more than twice as fast. Note, that flash attention works better with larger number of keypoints, hence there might not be significant differences when using only points.

We also highlight in \cref{fig:rotated_example} an example where GlueStick fails on a rotated image pair, while LightGlueStick manages to correctly match lines, even though it was not trained on such rotations. We hypothesize that the rotary encoding is a strong asset in this situation because of its translation equivariance that lets the network focus on modeling hard transformations like the rotation in \cref{fig:rotated_example}.

\begin{table}[h]
    \centering
    \resizebox{\linewidth}{!}{%
    \setlength{\tabcolsep}{4pt}
    \renewcommand{\arraystretch}{1.1}
    \small
    \begin{tabular}{l l c c c}
        \toprule
        \textbf{} & \textbf{Method} & \textbf{Point AP ($\uparrow$)} & \textbf{Line AP ($\uparrow$)} & \textbf{Time (ms) ($\downarrow$)} \\
        \midrule
        \multirow{4}{*}{\textbf{Points}} 
            & LightGlue (LG) ~\cite{lindenberger2023lightglue} & 76.4 & -- & {\bf 39} \\
            & GlueStick~\cite{pautrat_suarez_2023_gluestick} & {\bf 77.0} & -- & 92\\
            & \textbf{LightGlueStick} & {\bf 77.0} & -- & 44 \\
        \midrule
        \multirow{5}{*}{\makecell{\textbf{Points} \\ \textbf{+ Lines}}} 
            & L2D2~\cite{l2d2} & -- & 24.9 & 546 \\
            & LineTR~\cite{syoon_2021_linetr} & -- & 32.7 & 189 \\
            & SOLD2~\cite{pautrat_2021_sold2} & -- & 34.9 & 388 \\
            & LG+Endpoints & 76.7 & 58.6 & {\bf 38}\\
            & GlueStick~\cite{pautrat_suarez_2023_gluestick} & 76.5 & 70.9 & 106\\
            & \textbf{LightGlueStick} & {\bf 78.1} & {\bf 74.6} & 47 \\
        \bottomrule
    \end{tabular}
    }
    \caption{\textbf{Point and line matching evaluation on the ETH3D dataset~\cite{Schops_2017_eth3d}.} We report the point and line Average Precision (AP) in percentage, as well as the average execution time.}
    \label{tab:eth3d_evaluation}
\end{table}

\subsection{Homography Estimation on HPatches}

HPatches~\cite{hpatches_2017_cvpr} is a standard dataset used to benchmark homography estimation. It consists in 108 sequences displaying either homography viewpoint or illumination changes between image pairs. We match points and/or lines with the matchers to evaluate, use the 4-point homography solver implemented in PoseLib~\cite{PoseLib} when solving for the homography from points only, and use the same three minimal solvers combining 4 points, 2 points and 2 lines, or 4 lines in a hybrid RANSAC implementation~\cite{camposeco2018hybrid,Sattler2019Github} as in~\cite{pautrat_suarez_2023_gluestick} for the joint estimation. Using the ground truth homography, one can compute the precision and recall of the matching for both points and lines, as well as evaluate the quality of the retrieved homographies. For this, we compute the AUC of the reprojection error of the four image corners at three error thresholds: 1 / 3 / 5 pixels.

The results can be found in \cref{tab:homography_hpatches}. We compare separately point-only, line-only, and joint point-line methods. On points, the dense matchers~\cite{sun_2021_loftr,edstedt2024roma} obtain the best results, as could be expected from such powerful methods. However, these methods remain costly to use and are significantly slower than lightweight sparse matchers. For line matchers, LightGlueStick obtains the best precision and recall and is on par with GlueStick~\cite{pautrat_suarez_2023_gluestick} for homography estimation. The best performance is obtained after combining points and lines in the estimation, as the two features can complement each other depending on the scene. There, LightGlueStick scores again similarly or better compared to GlueStick~\cite{pautrat_suarez_2023_gluestick}, while being faster.

\begin{table}[t]
    \centering
    \resizebox{\linewidth}{!}{%
    \renewcommand{\arraystretch}{1.0}
    \small
    \setlength{\tabcolsep}{3.5pt}
    \begin{tabular}{llccc|cc|cc}
        \toprule
        & & \multicolumn{3}{c|}{\textbf{AUC (\(\uparrow\))}} & \multicolumn{2}{c|}{\textbf{Points (\(\uparrow\))}} & \multicolumn{2}{c}{\textbf{Lines (\(\uparrow\))}} \\
        \cmidrule(lr){3-5} \cmidrule(lr){6-7} \cmidrule(lr){8-9}
        & \textbf{Method} & 1px & 3px & 5px & P & R & P & R \\
        \midrule
        \multirow{5}{*}{\textbf{P}} 
        & RoMa & \textbf{44.0} & \textbf{72.3} & \textbf{81.3} & 95.8 & 97.0  & -- & -- \\
        & LoFTR & 40.6 & 69.1 & 78.5 & \textbf{98.2} & \textbf{99.6} & -- & -- \\
        \cmidrule(lr){2-9} 
        & LightGlue & 39.0 & 69.1 & 78.9 & 95.7 & 88.7 & -- & -- \\
        & GlueStick & 38.6 & 69.1 & 79.0 & 94.8 & 89.3 & -- & -- \\
        & \textbf{LightGlueStick} & 38.9 & 69.1 & 79.0 & 96.3 & 87.5 & -- & -- \\
        \midrule
        \multirow{5}{*}{\textbf{L}} 
        & L2D2 & 22.5 & 50.8 & 62.3 & -- & -- & 59.4 & 42.3 \\
        & LineTR & 15.6 & 42.2 & 54.2 & -- & -- & 81.1 & 53.8 \\
        & SOLD2 & 14.3 & 32.2 & 41.5 & -- & -- & 82.3 & 77.6 \\
        & GlueStick & \textbf{22.8} & \textbf{52.8} & 64.9 & -- & -- & 91.0 & 84.2 \\
        & \textbf{LightGlueStick} & 22.5 & 52.8 & \textbf{65.5} & -- & -- & \textbf{92.2} & \textbf{86.8} \\
        \midrule
        \multirow{3}{*}{\textbf{P+L}} 
        & PL-Loc & 35.7 & 63.6 & 74.1 & 86.3 & 68.3 & 80.9 & 53.5 \\
        & GlueStick & 38.5 & \textbf{69.9} & \textbf{79.9} & 95.3 & \textbf{89.4} & 90.4 & 86.7 \\
        & \textbf{LightGlueStick} & \textbf{39.2} & 69.8 & 79.4 & \textbf{97.3} & 87.2 & \textbf{92.7} & \textbf{87.2} \\
        \bottomrule
    \end{tabular}
    }
    \caption{\textbf{Homography estimation on HPatches~\cite{hpatches_2017_cvpr}.} We report the success rate AUC at 1 / 3 / 5 px thresholds, along with precision (P) and recall (R) for point and line matching.}
    \label{tab:homography_hpatches}
\end{table}

\subsection{Dominant Plane Estimation on ScanNet}
One of the standard applications combining points and lines is the estimation of a homography between two images. However, as noticed in previous works~\cite{pautrat_suarez_2023_gluestick}, the existing datasets used to benchmark homography estimation such as HPatches~\cite{hpatches_2017_cvpr} are already saturated, and a comparison on such datasets is moderately meaningful. Thus, we follow the example of these previous works~\cite{barath_23_homographies,pautrat_suarez_2023_gluestick}, in which an evaluation of homography estimation is proposed, by considering a pair of images, estimating the homography between the two dominant planes from feature matches, and then converting the retrieved homography into a relative pose~\cite{pose_from_H} and comparing it to the ground truth pose. We use the same solvers as in the previous experiment on HPatches to obtain the homography.

We reuse the 1500 image pairs selected in \cite{sarlin_2020_superglue}, originally taken from the ScanNet dataset~\cite{dai2017scannet}. These images come with the ground truth relative pose between the two images, making it possible to compare the homography-based relative pose with the ground truth one. Similarly to previous works~\cite{sarlin_2020_superglue,pautrat_suarez_2023_gluestick}, we report the pose error consisting of the maximum angular error in translation and rotation, as well as the pose AUC at 10\textdegree{} / 20\textdegree{} / 30\textdegree{} error thresholds.  For enhanced performance, the image resolution was reduced by half for LoFTR~\cite{sun_2021_loftr}. 

The results are displayed in \cref{tab:homography_scannet}. The dense matcher RoMa~\cite{edstedt2024roma} is the best, but at a prohibitive running time: it is more than 10 times slower than our method. For all three categories of sparse matched features, points-only, line-only, and points+lines, LightGlueStick performs the best. One can note that even though the point matching architecture of LightGlueStick remains similar to the one from LightGlue~\cite{lindenberger2023lightglue}, it can slightly improve upon the original LightGlue. The best results overall are obtained when combining points and lines, due to the complementary nature of the features: points excel in textured areas, while lines help on the frequent texture-less areas present in indoor data such as in ScanNet. Notably, LightGlueStick is able to improve the scores of GlueStick~\cite{pautrat_suarez_2023_gluestick} by 3 points in AUC.

\begin{table}[t]
    \centering
    \resizebox{\linewidth}{!}{%
    \setlength{\tabcolsep}{4pt}
    \renewcommand{\arraystretch}{1.1}
    \small
    \begin{tabular}{l l c c c}
        \toprule
        \textbf{} & \textbf{Method} & \textbf{Pose Error (\(\downarrow\))} & \textbf{Pose AUC (\(\uparrow\))} & {\bf Time (ms)} \\
        \midrule
        \multirow{4}{*}{\textbf{P}} 
        & RoMa~\cite{edstedt2024roma} & {\bf 6.77} & {\bf 34.8 / 54.7 / 65.6} & 593 \\
        & LoFTR~\cite{sun_2021_loftr} & 15.5 & 18.6 / 33.7 / 43.8 & 80\\
        \cmidrule(lr){2-5}
        & LightGlue~\cite{lindenberger2023lightglue} & 12.3 & 21.8 / 38.4 / 48.9 & {\bf 28} \\
        & GlueStick~\cite{pautrat_suarez_2023_gluestick} & 13.9 & 19.6 / 35.7 / 45.8 & 47\\
        & \textbf{LightGlueStick} & {\bf 12.0} & {\bf 22.5 / 39.1 / 49.3} & {\bf 28} \\
        \midrule
        \multirow{5}{*}{\textbf{L}} 
        & L2D2~\cite{l2d2} & 63.3 & 3.8 / 8.8 / 13.7 & 496\\
        & LineTR~\cite{syoon_2021_linetr} & 52.6 & 3.9 / 9.8 / 15.4 & 84\\
        & SOLD2~\cite{pautrat_2021_sold2} & 52.6 & 5.1 / 11.4 / 17.2 & 454\\
        & GlueStick~\cite{pautrat_suarez_2023_gluestick} & {\bf 27.0} & 9.5 / 19.9 / 29.0 & 60\\
        & \textbf{LightGlueStick} & 27.2 & \textbf{10.5 / 21.0 / 29.7} & {\bf 46}\\
        \midrule
        \multirow{3}{*}{\textbf{P + L}} 
        & PL-Loc~\cite{syoon_2021_linetr} & 24.2 & 12.1 / 24.6 / 33.3 & 169\\
        & GlueStick~\cite{pautrat_suarez_2023_gluestick} & 12.2 & 21.7 / 38.8 / 49.3 & 72\\
        & \textbf{LightGlueStick} & \textbf{10.8} & \textbf{24.3 / 42.0 / 52.5} & {\bf 49}\\
        \bottomrule
    \end{tabular}
    }
    \caption{\textbf{Relative pose estimation by dominant plane on ScanNet~\cite{dai2017scannet}}. We first estimate a homography based on point-only, line-only, or points+lines matches, then decompose it into the corresponding relative pose. We report the median pose error in degrees, as well as the AUC at 10\textdegree{} / 20\textdegree{} / 30\textdegree{} error.}
    \label{tab:homography_scannet}
\end{table}

\begin{figure*}[h]
    \centering
    \begin{tabular}{cc}
        \includegraphics[width=0.5\textwidth]{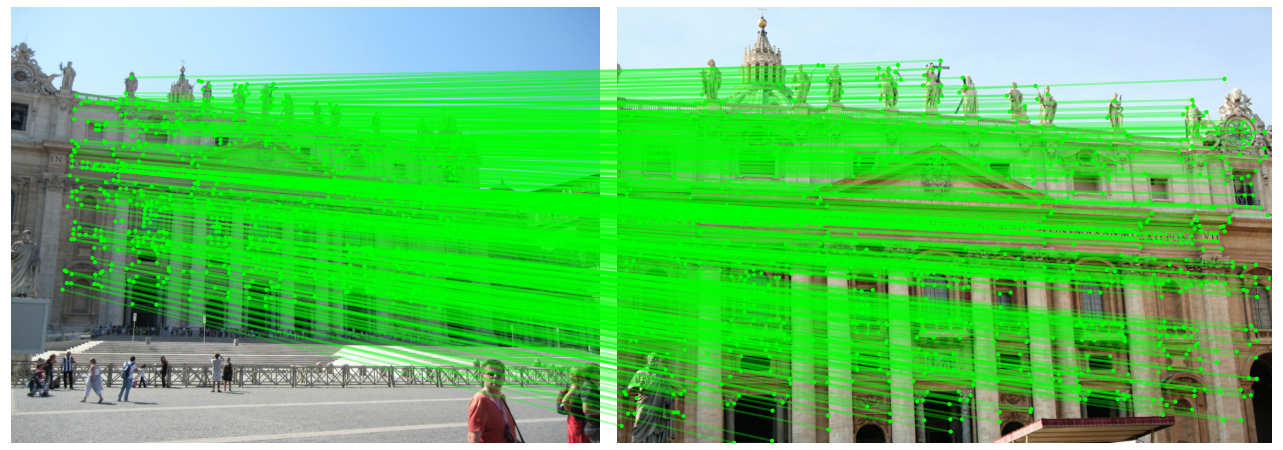} &
        \includegraphics[width=0.5\textwidth]{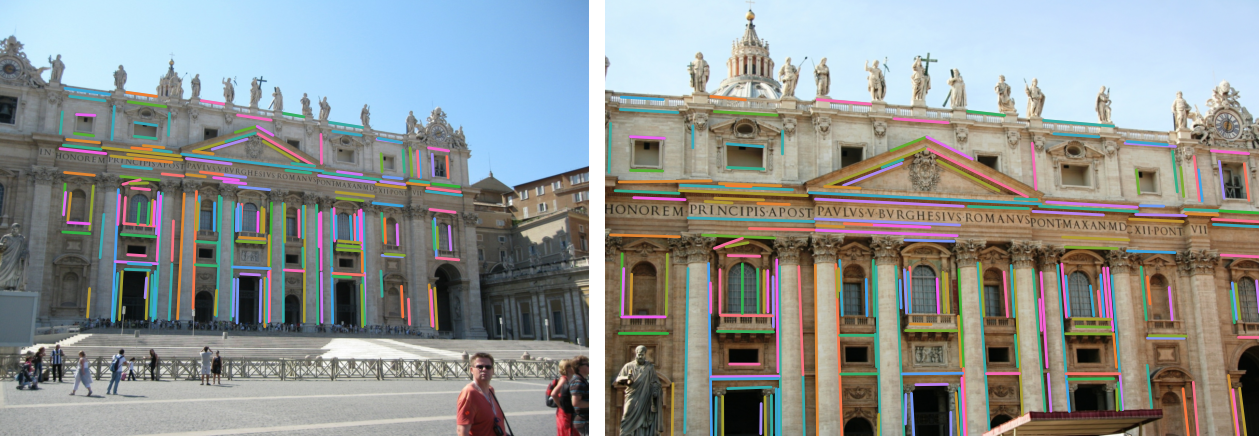} \\[5pt]
        \includegraphics[width=0.5\textwidth]{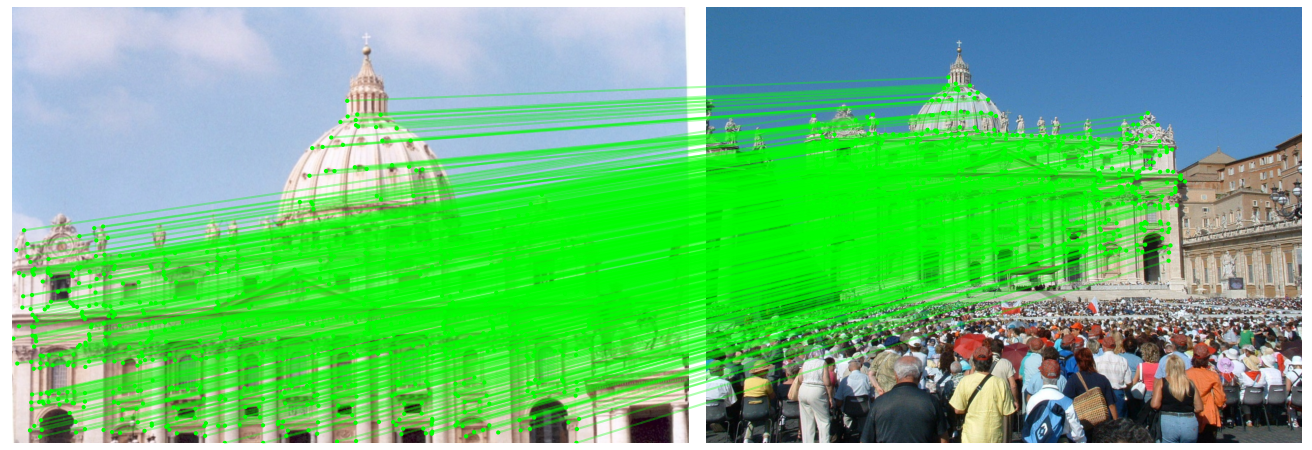} &
        \includegraphics[width=0.5\textwidth]{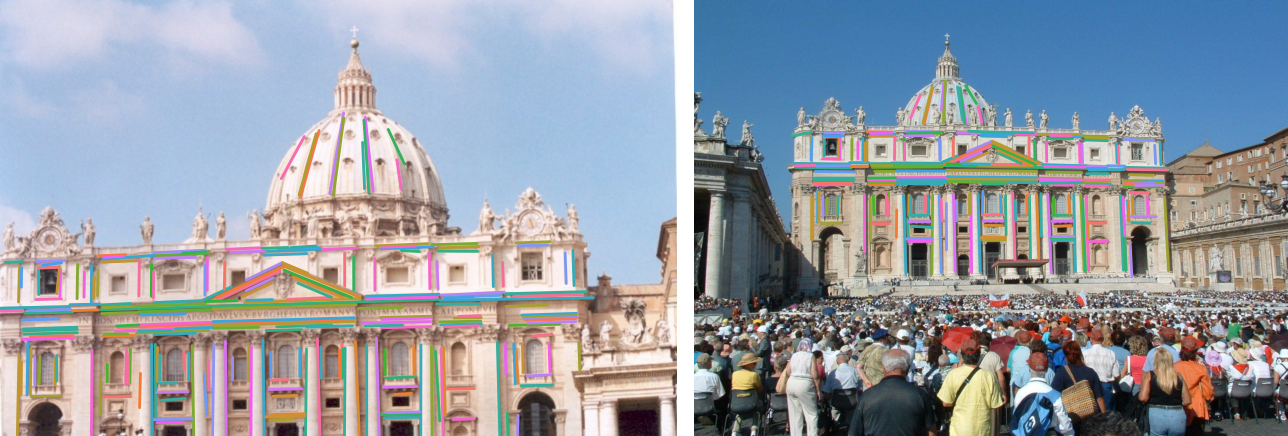} \\[5pt]
        \includegraphics[width=0.5\textwidth]{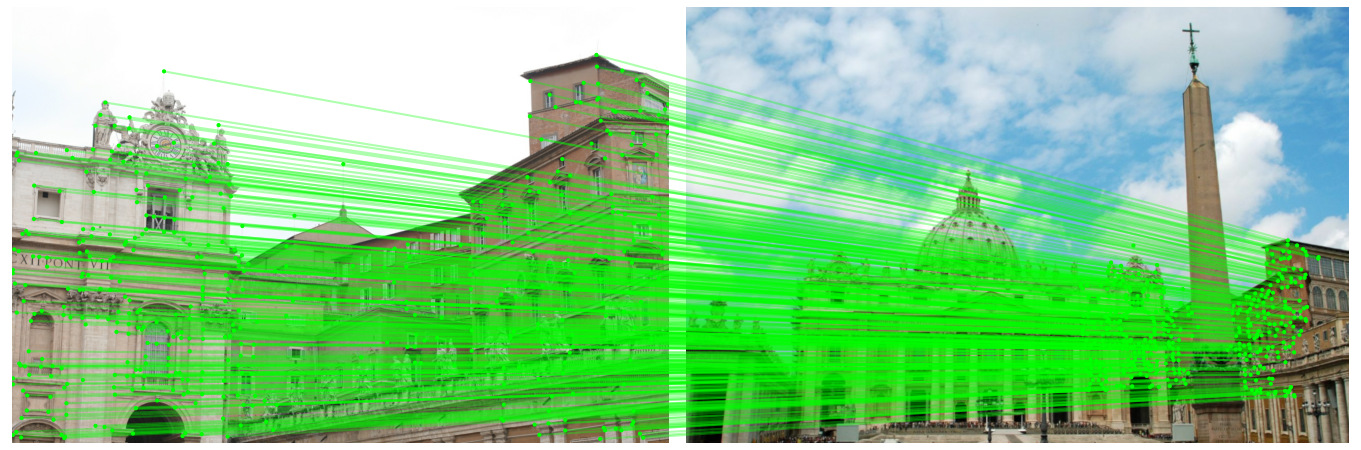} &
        \includegraphics[width=0.5\textwidth]{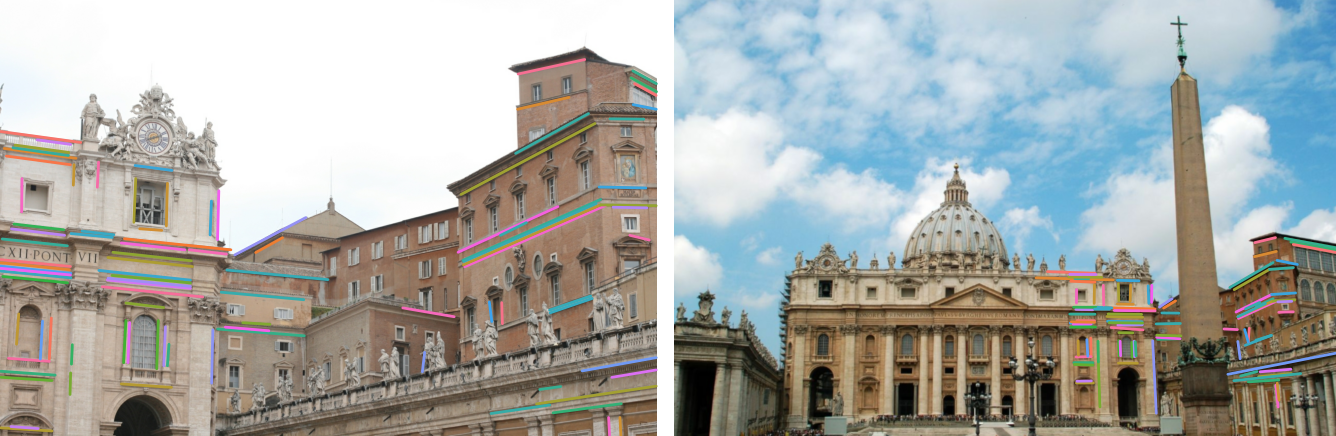} \\
    \end{tabular}
    \caption{{\bf Depth Adaptivity of LightGlueStick.} LightGlueStick adaptively adjusts its depth based on image difficulty, exiting after the 4th layer for the top pair, the 5th layer for the middle pair, and the 7th layer for the bottom pair. The bottom pair requires more layers due to its smaller visual overlap, making it more challenging to match. Processing times for the top, middle, and bottom pairs are 27ms, 34ms, and 42ms, respectively.}
    \label{fig:layer_viz}
\end{figure*}

\subsection{Visual Localization}

We also evaluate our joint point-line matcher on a visual localization task, where the goal is to find the camera pose of a query image with respect to a pre-built map coming from a set of database images. The 7Scenes dataset~\cite{7scenes} is a well-known dataset for this task, consisting of seven indoor scenes with RGB-D images and ground truth poses. We follow the hloc~\cite{sarlin_2020_superglue,hloc} and LIMAP~\cite{Liu_2023_LIMAP} frameworks by detecting SuperPoint feature points~\cite{detone2018}; extracting LSD lines~\cite{gioi_2010_lsd}; for every query image, retrieve the top 10 closest images in the database with NetVlad~\cite{arandjelovic2016netvlad}; and match them with the different matchers we are evaluating. When evaluating a pure line matcher, we use LightGlue~\cite{lindenberger2023lightglue} to match the points. Since ground truth depth is available, we use it to back-project lines to 3D on the database images, so that we get 2D-3D matches and can use the P3P solver of PoseLib~\cite{Ding_2023_CVPR,PoseLib} for points, and the joint point-line solvers of PoseLib~\cite{kukelova2016efficient,Zhou2018ASA,PoseLib} to estimate the absolute pose from points and lines. The 7Scenes is however quite saturated already, but the scene stairs remains the most challenging one, especially for point features due to the lack of texture and the repeated pattern of steps. Thus, we only show the results on this scene in \cref{tab:7scenes_stairs}.

As shown in the table, point-only methods perform similarly, but adding line features yields a notable boost due to their long range across the image and their presence even in texture-less areas. Among point+line methods, LightGlueStick matches the previous best, GlueStick~\cite{pautrat_suarez_2023_gluestick}, but is nearly four times faster: GlueStick processes 7.0 pairs per second, while LightGlueStick manages to process 26.9 pairs per second.

\begin{table}[ht]
\centering
\small
\setlength{\tabcolsep}{10pt}
\begin{tabular}{l c c c}
\toprule
\textbf{} & Method & T / R err. & Acc. \\
\midrule
\multirow{3}{*}{\makecell{\textbf{Points}}} & LightGlue~\cite{lindenberger2023lightglue} & 4.3 / 1.14 & 57.9 \\
& GlueStick~\cite{pautrat_suarez_2023_gluestick} & {\bf 4.1 / 1.08} & {\bf 60.3} \\
& LightGlueStick & {\bf 4.1} / 1.13 & 59.8 \\
\midrule
\multirow{5}{*}{\makecell{\textbf{Points} \\ \textbf{+ Lines}}} & SOLD2~\cite{pautrat_2021_sold2} & 3.1 / 0.82 & 78.1 \\
& LineTR~\cite{syoon_2021_linetr} & 3.1 / 0.81 & 76.8 \\
& L2D2~\cite{l2d2} & 2.9 / 0.77 & 78.4 \\
& GlueStick~\cite{pautrat_suarez_2023_gluestick} & {\bf 2.7 / 0.73} & 78.8 \\
& LightGlueStick & {\bf 2.7} / 0.74 & {\bf 79.1} \\
\bottomrule
\end{tabular}
\caption{\textbf{Visual localization on scene stairs of the 7Scenes dataset~\cite{7scenes}.} We report the median translation / rotation errors (cm / deg) and pose accuracy at 5cm / 5° threshold.}
\label{tab:7scenes_stairs}
\end{table}

%% file: sec/5_Conclusion.tex
\section{Conclusion}

In this work, we introduced a deep network for joint point and line matching that operates at near real-time speeds (20 FPS). Introducing novel architecture changes, we in particular explicitly encoded the line connectivity into the network, and let the network reason about non-repeatable features through a novel Attention Line Message Passing. We demonstrate that our method maintains or even improves upon the performance of state-of-the-art sparse matchers, while also being among the fastest point matchers available. Furthermore, our approach surpasses the fastest line matchers by a significant margin, setting a new benchmark for speed and accuracy. Additionally, by incorporating the adaptivity mechanism, we enable the network to predict line matches earlier, further reducing the runtime of our matcher, with only a marginal trade-off in performance. By enabling fast and accurate feature matching, our method unlocks the potential for real-time point-line feature fusion in embedded systems, such as robotics and mobile devices.

{\small 
\noindent{\bf Acknowledgments}\\
This work was supported by the strategic research project ELLIIT and the Swedish Research Council (grant no.~2023-05424).}